\documentclass{article}

\usepackage{arxiv}

\usepackage[utf8]{inputenc} 
\usepackage[T1]{fontenc}    
\usepackage{hyperref}       
\usepackage{url}            
\usepackage{booktabs}       
\usepackage{amsfonts}       
\usepackage{nicefrac}       
\usepackage{microtype}      
\usepackage{lipsum}        
\usepackage{graphicx}
\usepackage[numbers,compress]{natbib}
\usepackage{doi}

\title{Back to the Feature: Zero-Shot 6DoF Pose Estimation via
Dense Local Features}

\author{
    Ali Rafiaei \quad Michael Greenspan \\
    Dept. Electrical \& Computer Engineering, Ingenuity Labs \\
    Queen's University \\
    Kingston, ON, Canada \\
    \texttt{\{23rppc, michael.greenspan\}@queensu.ca}
}
\date{}

\renewcommand{\headeright}{}
\renewcommand{\undertitle}{}
\renewcommand{\shorttitle}{Rafiaei \& Greenspan: Back To The Feature}

\hypersetup{
    pdftitle={Back to the Feature: Zero-Shot 6DoF Pose Estimation via
    Dense Local Features},
    pdfsubject={6D Object Pose Estimation},
    pdfauthor={Ali Rafiaei, Michael Greenspan},
    pdfkeywords={6DoF Object Pose Estimation \and Novel Object Pose
        Estimation \and Zero-Shot Learning \and Foundation Models \and Dense
    Feature Matching \and Template Matching \and Training-Free},
}

\usepackage{amsmath}
\usepackage{amssymb}
\usepackage{xcolor}
\usepackage{colortbl}

\usepackage{graphicx}
\usepackage{subcaption}

\usepackage{pifont}
\usepackage{multirow}
\usepackage{booktabs}
\usepackage{arydshln}
\usepackage{todonotes}
\usepackage{geometry}
\usepackage{tabularx}
\usepackage{float}
\newcommand{\imgwidth}{0.21\linewidth}

\begin{document}
\maketitle

\begin{abstract}
    We present B2TFPose, a training-free zero-shot method for 6DoF pose
    estimation of unseen objects from RGB images.
    Using a single frozen DINOv3 vision transformer as its only pretrained
    component within the pose estimation pipeline, B2TFPose extracts dense
    patch-level features that generalize
    across the synthetic-to-real domain gap without any task-specific
    fine-tuning, revisiting the classical local feature matching paradigm
    through the lens of large-scale self-supervised foundation models.
    Three contributions advance the training-free state of the art. A
    geodesic non-maximum suppression strategy retrieves a
    viewpoint-diverse template set for coarse-to-fine correspondence
    matching. Render-guided Re-Correspondence (RRC) synthesizes
    object-specific views at the estimated pose and re-establishes dense
    2D-3D correspondences to sharpen the initial estimate without
    additional learned parameters. A multi-mask hypothesis selection
    strategy jointly scores competing segmentation candidates to
    resolve segmentation ambiguity.
    On the seven core datasets of the BOP Benchmark, B2TFPose achieves
    40.7 mean AR without refinement and 56.4 with refinement, establishing
    state-of-the-art performance among training-free RGB methods and
    outperforming trained counterparts including GigaPose and GenFlow,
    at competitive inference speed.
\end{abstract}


\section{Introduction}
\label{sec:intro}
Prior to the advent of deep learning,
progress toward 6 degree-of-freedom pose estimation (6DoF PE)
stemmed primarily from the effectiveness of classical local feature descriptors.
Starting with the introduction by Lowe of the
seminal Scale Invariant Feature Transform (SIFT)~\cite{lowe2004sift},
which led to a slew of variations and
alternatives~\cite{rublee2011orb,calonder2010brief,dalal2005hog,LI20081771},
these methods were all local in nature,
encoding mostly radiometric patterns in compact spatial
neighborhoods~\cite{tola2010daisy}.
They all aimed to maintain robustness to
in-plane rotations and scale~\cite{bay2006surf}, and variations in appearance
and lighting,
and while manually designed,
statistical methods were frequently employed to justify design
decisions and optimize hyperparameter operating
points~\cite{hartley2003mvg,mikolajczyk2005}.
Their resulting performance was impressive,
and nearest neighbor correspondences in feature space
could for the first time
establish spatial matches between query and target images,
enabling effective solutions to Autostitch~\cite{brown2007},
Structure-from-Motion~\cite{schoenberger2016sfm}, and object-level 6DoF PE
applications.

Following the success of AlexNet at image
classification~\cite{krizhevsky2012imagenet},
Machine Learning (ML) rapidly expanded to other pattern recognition-centric
computer vision tasks, including 6DoF PE.
The classical-era labelled LINEMOD dataset~\cite{linemod} was
repurposed for supervised 6DoF PE
workflows~\cite{GDRNet, PVNet},
and
YCB-Video~\cite{ycbv}
and other purpose-built annotated datasets soon followed,
resulting in the rapid advance of
fully supervised 6DoF PE methods~\cite{BOP}.
The performance of these emerging ML-based methods definitively surpassed
that of the preceding classical approaches~\cite{BOP},
and research focus on local features diminished.

Despite this progress, something significant was lost in the
transition from the classical to the ML era.
Classical feature descriptors were inherently local, robust, and
general~\cite{schoenberger2017}: once defined, local features could effectively
encode previously unseen objects
without training or even tuning.
In this way, classical methods were
intentionally designed to work with unseen objects, i.e. they were
inherently ``zero-shot''.
In contrast, learned methods tended to be global and object-specific.
While they worked remarkably well
on objects represented in the training dataset, they performed much less
effectively (or not at all) for previously unseen
``out of distribution'' objects.
This limitation is partly attributable to network architectures
that, in deeper layers, encode large receptive fields that
encompass the entirety of an object~\cite{luo2017erf},
effectively yielding global features.
Whereas learned methods show leading performance when
full supervision is possible, they falter when encountering the
zero-shot case of unseen objects.
The main question then that this work addresses is:

\vspace{7pt}
\textit{Is there an approach to 6DoF PE that combines
    the superior performance of supervised machine learning,
    with the zero-shot capabilities
    inherent
to classical local feature descriptors?}
\vspace{7pt}

The answer to this question involves building on advances in
foundational vision transformers (ViTs)~\cite{vit}.
Foundation models such as DINOv3~\cite{DINOv3}
and CLIP~\cite{clip} are
trained on large-scale datasets and therefore generalize well to previously
unseen content.
Further, the very architecture of vision transformers,
wherein input images are tokenized into a grid of
small (typically $16\!\times\!16$ pixel) patches,
naturally leads to the encoding of limited receptive fields~\cite{raghu2021}.
The handling of these encoded patches then becomes analogous to the
matching pipeline established in classical approaches,
which benefits further from the advent of effective zero-shot
instance segmentation~\cite{CNOS, SAM}.
A few works have been recently introduced that follow this approach,
such as FoundPose~\cite{FoundPose} and
FreeZe~\cite{freeze},
which start with zero-shot segmentation, extract DINO features,
find correspondences in feature space,
and estimate and refine pose.
While each method introduces alternatives
in the matching, estimation and refinement steps,
they all make use of foundational local features
and follow the same basic canonical processing pipeline introduced
by Lowe~\cite{lowe2004sift}.

In this work, we focus on improving three particular steps of this
pipeline: coarse feature matching, pose refinement, and mask
hypothesis selection.
Coarse matching narrows the correspondence search to a subset of
view-sphere templates before fine matching; FoundPose~\cite{FoundPose}
retrieves only 5 templates via bag-of-words retrieval, while
FreeZe~\cite{freeze} bypasses coarse matching entirely.
We show instead that the number of templates admitted at this stage
significantly impacts both the effectiveness and efficiency of fine
matching and pose recovery.

Pose refinement addresses residual error in the initial PnP estimate
arising from imperfect correspondences, occlusion, or rotational
ambiguity.
Existing training-free methods refine pose via render-and-compare
against the initial estimate, without regenerating 2D-3D
correspondences for a second geometry solve.
We introduce \emph{Render-guided Re-Correspondence} (RRC): the object
is rendered at the initial pose estimate and a fixed set of rotation
variants, and a second round of dense 2D-3D correspondences is
established between the query and these renders using the same frozen
DINOv3 backbone, substantially improving pose accuracy without
additional learned parameters.

The final selection stage must identify the best pose hypothesis when
multiple competing candidates arise.
Zero-shot instance segmentation methods such as CNOS~\cite{CNOS} return
several candidate masks at varying confidence levels, and most prior
approaches reduce these to a single top-confidence mask prior to pose
estimation, effectively discarding alternatives that may more accurately
delineate the object region.
We instead retain the top candidate masks and run the
full correspondence and pose estimation pipeline independently for each,
yielding a richer hypothesis pool, from which the best can be identified
by scoring each hypothesis with a combination of visual appearance
similarity, mask overlap with the projected model, and segmentation
confidence.

A significant aspect of our approach is that, like classical methods,
it is training-free. While we do make use of DINOv3~\cite{DINOv3},
which is a pre-trained foundational transformer, we do not rely on
any additional fine-tuning or training of other network components.
This attribute is highly desirable in deployments of zero-shot 6DoF PE
where training is not practical,
such as augmented reality and industrial part handling.
The novel contributions of this work are:
\begin{itemize}

    \item A geodesic-NMS template retrieval strategy for coarse
        correspondence matching that selects
        $k$ diverse viewpoints, improving pose recovery over both
        few-template and exhaustive alternatives, as validated through
        ablation.

    \item A novel Render-guided Re-Correspondence (RRC) pose refinement
        step that synthesizes
        object-specific views at the estimated pose to establish a second
        round of correspondences, substantially improving pose accuracy
        without additional learned components.

    \item A multi-mask hypothesis selection strategy that retains
        competing segmentation candidates and scores the resulting pose
        hypotheses, improving robustness to segmentation ambiguity.

    \item B2TFPose, a training-free zero-shot pipeline built on a
        single frozen DINOv3~\cite{DINOv3} model, achieving
        state-of-the-art performance among training-free RGB methods
        while running faster than the closest comparable baseline.

\end{itemize}

\section{Related Work}
\subsection{Instance-Level Pose Estimation}

Early 6DoF pose estimators relied on hand-crafted feature templates.
Hinterstoisser~\cite{linemod} introduced LINEMOD, encoding
gradient orientations and
surface normals into compact multi-modal templates rendered from
object CAD models, demonstrating
real-time detection of texture-less objects in cluttered scenes.
These classical approaches generalize poorly to large viewpoint
change, heavy occlusion, and
appearance variation, motivating the shift to learned representations.

The emergence of deep convolutional networks transformed the field.
PoseCNN~\cite{ycbv} pioneered direct regression of object pose from RGB images,
while SSD-6D~\cite{SSD6D} extended single-shot object detection to
cover the full rotation space.
PVNet~\cite{PVNet} proposed pixel-wise voting over learned direction
vectors pointing to
keypoints, providing robustness to partial occlusion by aggregating
cues from all visible
surface patches.
RCVPose~\cite{wu2022rcvpose} replaces directional voting with a radial
scheme, in which each point casts a vote via an intersecting sphere
toward a small set of dispersed keypoints, improving accuracy under
heavy occlusion; RCVPose3D~\cite{wu2022rcvpose3d} extends this radial
voting formulation to point-cloud-only input, separating semantic
segmentation from keypoint regression.
The Augmented Autoencoder~\cite{AAE} removed the need for real
annotated images entirely by
training on domain-randomized synthetic renders and recovering pose
via nearest-neighbor lookup
in a learned embedding space.

Later work shifted toward dense surface correspondence representations.
GDR-Net~\cite{GDRNet} predicts geometry-guided dense correspondence
maps and recovers pose via
a differentiable PnP layer, while ZebraPose~\cite{ZebraPose} encodes
object surface coordinates
in a hierarchical binary scheme enabling coarse-to-fine disambiguation.
CosyPose~\cite{CosyPose} extended single-object estimation to
multi-object, multi-view scenes
with a render-and-compare refiner, achieving leading results on the
BOP benchmark~\cite{BOP}
for several years.
Efforts to generalize within a semantic category~\cite{NOCS} reduce
but do not eliminate the
per-object data collection burden.
In all of these approaches, a separate training phase for each object
or object category is
required, fundamentally limiting deployment to objects seen at training time.
Even self-supervised variants that reduce reliance on real annotations,
such as RKHSPose~\cite{wu2024rkhspose}, which adapts a synthetically
pretrained keypoint-voting network to real images using pseudo-labels
rather than ground-truth pose, still require this per-object training
phase.

\subsection{Zero-Shot Object Pose Estimation}

Zero-shot object pose estimation targets the challenging setting in which
objects are entirely unseen
at training time and only a CAD model is available at test time.
MegaPose~\cite{MegaPose} established the first scalable learned solution by
training a coarse pose
estimator and render-and-compare refiner on a large-scale synthetic
dataset spanning thousands
of diverse objects, demonstrating that breadth of training objects
enables generalization to
novel instances.
TemplatePose~\cite{TemplatePose} showed that template matching with
learned local features can
likewise generalize to new objects without per-object retraining,
serving as an early bridge
between the supervised and zero-shot regimes.

Subsequent methods improved accuracy, speed, and generality.
GigaPose~\cite{GigaPose} decomposes pose search into out-of-plane
recovery from discriminative
rendered templates followed by in-plane estimation via patch
correspondences, achieving a
35$\times$ speedup over prior works.
GenFlow~\cite{GenFlow} iteratively refines pose by predicting optical
flow between rendered and
observed images under 3D shape constraints in a differentiable pipeline.
Co-op~\cite{coop} advances the correspondence paradigm with
semi-dense matching and a
probabilistic flow model, achieving the highest RGB-only accuracy on the BOP
benchmark~\cite{BOP} among zero-shot methods, at the cost of
requiring large-scale
synthetic training data.
SAM-6D~\cite{sam6d} pairs the Segment Anything Model~\cite{SAM} with
point-cloud descriptor
matching for zero-shot instance detection and pose estimation,
requiring depth as input.
Despite their strong performance, all of these methods depend on
large-scale synthetic training
pipelines, incurring significant data curation costs and limiting
applicability to settings
where training is impractical.

Training-free methods take a different route, exploiting features
from large-scale pretrained foundation models that generalize
across the synthetic-to-real
domain gap without task-specific adaptation.
FoundPose~\cite{FoundPose} retrieves candidate templates via a
bag-of-words index over
DINOv2~\cite{DINOv2} patch descriptors, establishes dense 2D-3D
correspondences by
nearest-neighbor matching, and refines the PnP estimate with
featuremetric alignment and a
render-and-compare refiner.
ZS6D~\cite{zs6d} similarly retrieves pose via template similarity
using a self-supervised ViT, but selects only the single
highest-scoring template per query, without an explicit diversity
mechanism among candidates.
Pos3R~\cite{deng2025pos3r} instead builds on the pretrained 3D
foundation model MASt3R~\cite{leroy2024mast3r} to establish
correspondences between the query and only forty rendered templates,
selecting the best template by correspondence quality rather than a
separate retrieval step.
FreeZe~\cite{freeze} (RGB-D) combines 3D geometric point descriptors
from GeDi~\cite{gedi}
with DINOv2 visual features to form discriminative point-level
representations, recovering
pose by 3D-to-3D RANSAC registration. Success of these training-free
methods, even over methods that require training, confirms that frozen
foundation model features are sufficient for competitive zero-shot pose
estimation without any training on task-specific data.

\subsection{Foundation Models for Visual Feature Extraction}

The Vision Transformer (ViT)~\cite{vit} processes images as sequences
of fixed-size patch
tokens, each encoding a small spatial neighborhood.
This patch-level structure naturally limits the effective receptive
field of individual
tokens~\cite{raghu2021}, producing spatially grounded local representations that
contrast sharply with
deep CNN features, which aggregate context across large image
regions~\cite{luo2017erf}.
With the introduction of DINOv1~\cite{DINOv1},
it was shown
that training ViTs with
self-distillation and enforcing
consistent representations across different augmented crops of the
same image without any
human labels, yields patch tokens with strong cross-image semantic
correspondence
properties.
DINOv2~\cite{DINOv2} scaled this approach with curated training data
and larger model
capacities, producing all-purpose visual features validated across
dense prediction tasks
including depth estimation, segmentation, and cross-image
correspondence. It is the visual feature
backbone that is used by FoundPose~\cite{FoundPose} and FreeZe~\cite{freeze}.
DINOv3~\cite{DINOv3} advances the family further with improved
generalization and
representational capacity; it serves as the sole learned component in B2TFPose.

Alongside the DINO family, CLIP~\cite{clip} demonstrated that
contrastive image-text
pretraining on web-scale data produces visual features that transfer
broadly across
recognition tasks without fine-tuning.
SAM~\cite{SAM} extended the foundation
model paradigm to
instance segmentation, enabling high-quality mask generation for
arbitrary objects from
minimal user prompts.
CNOS~\cite{CNOS} combines these advances for zero-shot object detection
and generates mask
proposals using a fast SAM variant and scores them against multi-view
object templates using
DINOv2 features, enabling zero-shot instance segmentation from a CAD
model alone.
CNOS serves as the detection front-end for B2TFPose and all competing
methods in our
evaluation, ensuring that performance differences reflect the pose
estimation pipeline rather
than the object detector.

\section{Method}

\subsection{Overview}
\begin{figure}[t]
    \centering
    \includegraphics[width=\textwidth]{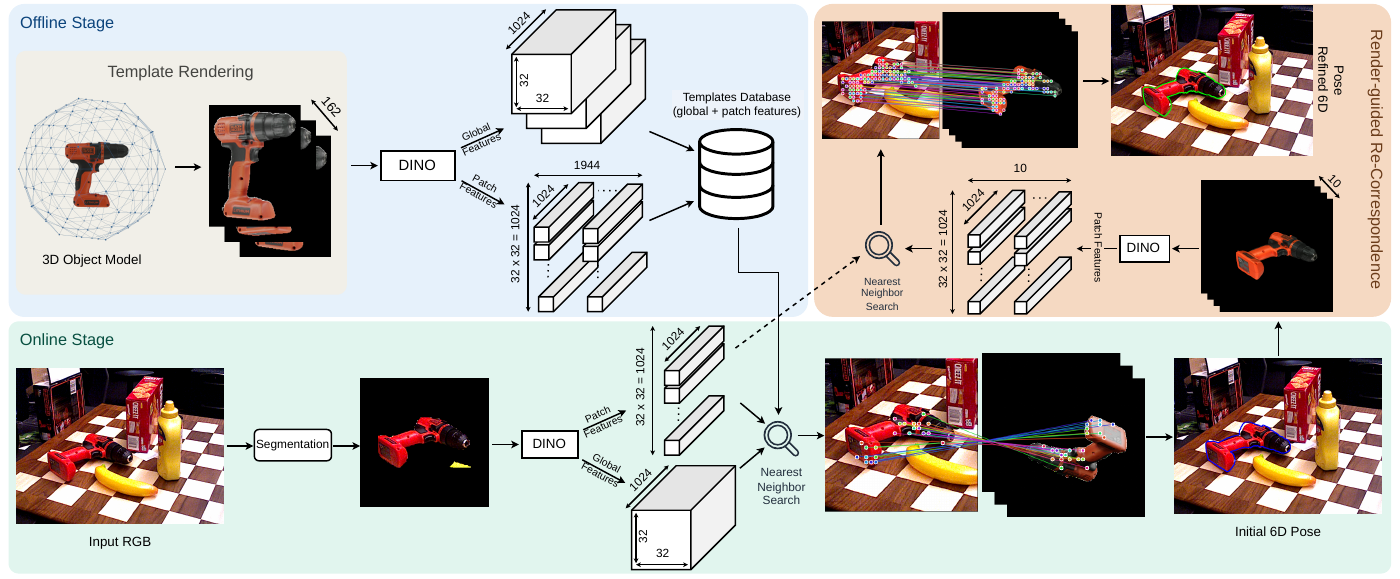}
    \vspace{1pt}
    \caption{\textbf{Overview of B2TFPose.} \textit{Offline stage}
        (top-left): the 3D
        object model is rendered from 162 uniformly distributed viewpoints and
        augmented with 12 in-plane rotations to yield 1944 templates, each
        encoded by a frozen DINOv3 backbone into a global 1024-D descriptor for
        coarse retrieval and a $32{\times}32$ grid of local patch descriptors for
        dense matching; both are stored in a template database.
        \textit{Online stage} (bottom): the input RGB image is segmented and encoded
        by the same DINOv3 backbone; global feature similarity retrieves the most
        viewpoint-diverse candidate templates from the database, and mutual
        nearest-neighbor patch matching establishes 2D-3D correspondences that
        are passed to RANSAC+PnP for an initial 6DoF pose estimate.
        \textit{Render-guided Re-Correspondence} (top-right): ten
        views are rendered at orientation variants around the initial estimate,
        re-encoded by DINOv3, and re-matched against the query patch features
        to produce a second, tighter set of 2D-3D correspondences; a
        final RANSAC+PnP
        yields the refined pose.
    }
    \label{fig:overview}
\end{figure}

Given a query RGB image and the 3D mesh model of an unseen rigid
object, our method,
which we call ``Back to The Feature'' or \textit{B2TFPose},
estimates the object's 6DoF pose zero-shot, i.e. without any
task-specific training. Our
method builds on the notion that the dense spatial features produced
by a pretrained Vision
Transformer (ViT) encode strong local appearance descriptors that
generalize well across the synthetic-to-real domain gap.
B2TFPose operates in two stages, as illustrated in Fig.~\ref{fig:overview}.
The \emph{offline stage} is run once for every new object and includes
template generation and encoding of the rendered views of the object
mesh into a searchable database.
This onboarding stage is
limited to 5 minutes of processing time and utilization of only one
GPU as specified by the BOP challenge guidelines~\cite{BOP}.
The \emph{online stage}, which runs per image, proceeds with
the following steps: the
query object in the image is first segmented using a zero-shot
segmentation method,
resulting in a center-cropped image of the object.
Following segmentation, a coarse-to-fine strategy initially retrieves
a set of template views
that are visually similar to the cropped image, followed by
establishing dense 2D-3D point correspondences between the query image and the
retrieved templates to estimate and refine the object pose.
The pose estimation pipeline uses a single frozen
DINOv3~\cite{DINOv3} model as its only
pre-trained component, requiring no fine-tuning or object-specific adaptation.

\subsection{Offline Template Generation and Encoding}

We follow a common practice in template-based pose estimation
methods~\cite{TemplatePose,
MegaPose, GigaPose} and pre-render object templates from a set of uniformly
distributed viewpoints on the view sphere. Specifically, we
approximate uniform sampling using a
discretization based on a level 2 icosahedron, yielding 162 rendered
views per object, as in
TemplatePose~\cite{TemplatePose} and GigaPose~\cite{GigaPose}.
For each viewpoint, we render an RGB image, a segmentation mask, and
a depth map, using
fixed camera intrinsics via BlenderProc~\cite{BlenderProc}. The
associated poses of all
templates are recorded and used in later steps to establish an
image-frame point in
object-frame coordinates. We refer to these 162 rendered images as
\emph{base views}.
Upon generation, the in-plane rotation of each base view is
arbitrary. To cover the full
$360^\circ$ of in-plane rotations, each base view is augmented with
$12$ discrete in-plane
rotations at $30^\circ$ intervals, yielding $162 \times 12 = 1944$
templates in total.
Each of the 1944 templates is encoded into a compact global template
descriptor for coarse
retrieval. The template is  centered on the object and cropped to a
square bounding box with
padding and resized to $512 \times 512$ pixels. It is passed through
DINOv3 from which the final
layer's patch tokens are extracted, resulting in a $32 \times 32$
grid of $1024D$ tokens.
These tokens are then mean-pooled over foreground patches only,
those whose corresponding
downsampled mask value is non-zero, yielding a single  $1024D$
global descriptor per template
that is robust to segmentation noise and partial occlusion.

In addition to global template descriptors, at this stage we also
encode local patch-level features.
We run a single DINOv3 forward pass on each cropped template image
and extract the
foreground patch embeddings together with their corresponding pixel centers
and sampled depth values.
There are two differences from the template view global feature
extraction, the first
being that, rather than averaging the features into a single
descriptor, each patch
maintains its own distinct $1024D$ descriptor.
The second difference is that the features here are  extracted from
an intermediate
transformer block layer, which have been shown to provide a better
balance of semantic
and spatial information than the final layer~\cite{FoundPose}.
Only patches whose corresponding location in the downsampled mask is
foreground are
retained.
These centers are backprojected into the object model frame using the
rendered depth
and the known template pose, so that no backprojection is required
during the online
stage at query time. The cache is built once during the offline stage
and is reused
across all queries for a given object.
The output of the offline stage is therefore:
\begin{itemize}
    \item A set of 162 rendered base views per object, each with an RGB
        image, depth
        map, segmentation mask, and associated pose.
    \item A database of 1944 global template descriptors, one per view,
        computed by
        masked average pooling of DINOv3 patch tokens over foreground patches.
    \item A database of dense patch features for all 1944 templates,
        where each entry
        includes the foreground patch embedding and its pre-computed 3D
        object-frame coordinates of the patch center.
\end{itemize}

\subsection{Online Stage}
The online stage comprises five sequential steps: segmentation,
coarse view retrieval,
dense matching and pose estimation, refinement, and hypothesis selection.

\noindent\textbf{Segmentation.}
We obtain segmentation mask proposals for the target object by
applying CNOS~\cite{CNOS},
a zero-shot segmentation method, to the query RGB image.  CNOS
returns an instance segmentation,
along with an initial classification and confidence score for each segment,
based on visual similarity with object templates rendered by CNOS.
It has been observed~\cite{freeze} that in some cases, the
highest-scoring candidate is
not always correct, so we retain up to two candidates that
exceed a confidence threshold of 0.35 (chosen empirically),
falling back to one if fewer qualify.
Considering more candidates would improve robustness at the cost of
proportionally
higher runtime, making two a practical choice.
The subsequent steps including coarse view retrieval, dense matching, and
refinement, are executed independently for each retained mask.

\noindent\textbf{View Retrieval with Global Features.}
For each retained segmentation mask, we encode the query RGB image using the
same procedure as for the template global descriptors.
We then retrieve the top $k$ template candidates by cosine similarity
(i.e. nearness to the query embedding) while enforcing a minimum
geodesic rotational
distance of $\theta\!=\!20^\circ$
between any two selected views, through non-maximum suppression.
Non-maximum suppression operates on the full combined rotation of each
template (out-of-plane viewpoint and in-plane rotation together).
This ensures the selected candidate set spans a broad range of viewpoints rather
than being limited to a similar and small range of template poses,
improving robustness for ambiguous, occluded or symmetric objects.
Through hyperparameter tuning experiments we found $k=125$ to work well.

\noindent\textbf{Patch Matching with Local Features.}
For each query image, we extract local patch-level features using the
same procedure
described above for the offline stage, resulting in a $32 \times 32$
grid of $1024$-dimensional
query image features.
For each of the query foreground patches,
we establish a correspondence to a template patch using a greedy approach.
We first retrieve the top-3 nearest template
patch descriptors using nearest neighbor search across the pooled
patches of all retrieved templates.
To remove ambiguous matches,
we then apply Mutual Nearest Neighbor (MNN) filtering
in which a match between query patch $i$ and template patch $j$
is retained only if $j$'s nearest neighbor in the query set is also
$i$.
The bidirectional consistency check removes one-to-many matches and
reduces the fraction of outlier correspondences entering RANSAC.

For each retained match between a query patch and a template patch,
we establish a
2D-to-3D correspondence. The query patch center is taken as the 2D point and the
corresponding template
3D point on the object surface is read directly from the cache, where
it has already been
backprojected into the object model frame during the offline stage.
The pair constitutes a 2D-to-3D correspondence between an observed
location in the scene
(camera frame) and a point on the object surface (object frame).
From the set of three or more 2D-3D correspondences, we estimate the
rigid transformation using RANSAC$+$PnP~\cite{PnP}.

\noindent\textbf{Render-guided Re-Correspondence (RRC).}
\label{sec:method-rrc}
The initial RANSAC$+$PnP estimate provides a coarse pose, but the
correspondences that produced it come from the offline template library,
whose rendered views may differ substantially in appearance from the
query at the estimated viewpoint.
RRC exploits the initial estimate to generate a set of rendered views
that are far closer in appearance to the query, uses them to establish
a second round of correspondences, and runs another RANSAC$+$PnP to
refine the pose.

Specifically, given the initial pose $\hat{\mathbf{T}} = [\hat{\mathbf{R}}
\mid \hat{\mathbf{t}}]$, we construct ten pose variants by
right-multiplying the rotation with ten axis-aligned rotation deltas
$\Delta R_i \in \{I, R_x^{90}, R_x^{180}, R_x^{270}, R_y^{90}, \ldots\}$
(the identity plus nine $90^\circ$/$180^\circ$/$270^\circ$ increments
about each canonical axis), keeping the translation fixed, i.e.
$\mathbf{T}_i =
\begin{bmatrix} \hat{\mathbf{R}}\,\Delta R_i & \hat{\mathbf{t}}
\end{bmatrix}, \;i = 1\ldots 10
$.
Right-multiplying applies each rotation in the \emph{object} frame,
yielding variants that cover the principal orientations around the
estimated pose.
Axis-aligned increments at regular $90^\circ$ intervals are chosen for
their simplicity and systematic coverage of the rotation space without
additional hyperparameter tuning.
This targets a common failure mode in which near-symmetric or
repeating-part geometry causes the initial PnP solve to settle on a
plausible but rotationally offset pose; unlike widening the coarse
retrieval count $k$, which would sample generic additional viewpoints,
these renders are anchored to the current estimate's translation and
scale, testing axis-aligned alternatives to the specific hypothesis
already in hand.
The object mesh is rendered at all ten poses using the query camera
intrinsics, and every valid render (non-empty foreground mask) is
cropped and passed through DINOv3 (the same intermediate layer used
for patch matching) in a single batched forward pass.
For each render, foreground patch embeddings are extracted, and each
patch center is backprojected to 3D in the object frame using the
rendered depth map and pose $\mathbf{T}_i$.
The embeddings and 3D coordinates from all renders are pooled into one
combined set $\mathcal{E}^{\mathrm{rrc}}$.

Query foreground patch embeddings are then matched against
$\mathcal{E}^{\mathrm{rrc}}$ via nearest-neighbor search in feature
space, and the
resulting 2D-3D correspondences are passed to a single RANSAC$+$PnP run
to produce the RRC-refined pose.
The RRC-refined pose is then passed to MegaPose refinement~\cite{MegaPose} for
final refinement.

\noindent\textbf{Mask Hypothesis Selection.}
\label{para:mask-selection}
Each retained CNOS mask candidate is processed independently through
coarse view retrieval, patch matching, RANSAC$+$PnP, RRC, and MegaPose refinement,
yielding one refined pose per mask.
The final step selects from these competing poses.
The object is rendered at each candidate pose and hypotheses whose
rendered mask has an IoU below 0.2 with the corresponding
CNOS segmentation mask are discarded.
Each surviving hypothesis is then scored by three complementary signals:
\begin{itemize}
    \item $s_\text{cos}$: DINOv3 cosine similarity between the global
        descriptors of the rendered and observed object crops, measuring
        \emph{appearance alignment}.
    \item $s_\text{IoU}$: IoU between the rendered mask and the CNOS
        segmentation mask, measuring \emph{silhouette alignment}.
    \item $s_\text{cnos}$: CNOS detection confidence for the
        corresponding mask, encoding the segmentation model's certainty that
        the mask covers the target object.
\end{itemize}
Because the three signals live on different scales, meaning cosine similarity
is confined to a narrow range, whereas IoU and CNOS scores span a wider
range with different variances, each is z-score normalized across the
competing hypotheses before combining.
The final score is a weighted sum of the normalized signals:
\begin{equation}
    s = w_\text{cos}\,\hat{s}_\text{cos}
    + w_\text{IoU}\,\hat{s}_\text{IoU}
    + w_\text{cnos}\,\hat{s}_\text{cnos},
    \label{eq:score}
\end{equation}
where $\hat{\cdot}$ denotes the z-normalized score and the weights
$(w_\text{cos}, w_\text{IoU}, w_\text{cnos}) = (0.2, 0.5, 0.3)$ were
determined empirically.
The hypothesis with the highest combined score is returned as the final
pose estimate.

\section{Experiments}
\label{sec:experiments}

\subsection{Experimental Setup}

\noindent\textbf{Datasets.}
We evaluate on the seven core datasets of the BOP Benchmark~\cite{BOP}:
LM-O~\cite{lmo} (occlusion-heavy tabletop objects),
T-LESS~\cite{tless} (textureless industrial parts),
TUD-L~\cite{BOP} (objects under changing illumination),
IC-BIN~\cite{icbin} (bin-picking with clutter),
ITODD~\cite{itodd} (dark, near-textureless industrial objects),
HB~\cite{hb} (household objects in two scenes),
and YCB-V~\cite{ycbv} (everyday objects with heavy occlusion).
All datasets provide RGB-D channels, although we use only the RGB channels for
our results.

\noindent\textbf{Evaluation Metric.}
We report the BOP Average Recall (AR), the official BOP metric~\cite{BOP}.
A pose estimate is considered correct if its error falls below a threshold
$\tau$, and recall is the fraction of object instances in a scene for which a
correct estimate is returned.
AR averages this recall across multiple values of $\tau$ and across three
complementary error functions: the Visible Surface Discrepancy (VSD), which
measures the depth discrepancy between the rendered object at the estimated and
ground-truth poses over co-visible surface regions; the Maximum
Symmetry-aware Surface Distance (MSSD), which measures the maximum
point-to-point
distance between object surfaces, taking symmetries into account; and
the Maximum
Symmetry-aware Projection Distance (MSPD), which applies the same logic in 2D
image space.
Together, the three functions probe geometric accuracy in depth, 3D surface
alignment, and image-plane projection respectively.

\noindent\textbf{Detection and Segmentation.}
Following standard practice~\cite{FoundPose, MegaPose, GigaPose,
GenFlow, coop} and in
accordance with BOP benchmark guidelines~\cite{BOP}, we use CNOS~\cite{CNOS} as
the zero-shot instance segmentation method as do all methods in our
comparison,
ensuring that differences in the results reflect the pose estimation pipeline
rather than object detection and segmentation.

\noindent\textbf{Implementation Details.}
B2TFPose uses a single frozen DINOv3 ViT-L/16 model throughout.
Global template descriptors are extracted from the final transformer
block (layer 24),
while local patch correspondences use features from an intermediate
block (layer 18 of 24),
following the observation of~\cite{FoundPose} that intermediate
layers offer a better
balance of semantic and spatial detail.
Templates are rendered at 162 viewpoints on a level-2 icosahedron and augmented
with 12 discrete in-plane rotations at $30^\circ$ intervals, giving
$1944$ templates per object.
Coarse retrieval selects $k\!=\!125$ diverse templates using geodesic
non-maximum
suppression with a minimum inter-view angle of $\theta\!=\!20^\circ$;
the effect of $k$
is studied in Sec.~\ref{sec:ablation}.
Per query patch, the top-3 nearest template patches
from among up to $1024 \times\!125= 128{,}000$ foreground patches
from the 125 templates
are retrieved via
FAISS~\cite{faiss}
and filtered by mutual nearest-neighbor (MNN) consistency.
From the corresponding query and template patches, pose is then
estimated by RANSAC+PnP with a reprojection threshold of 4 pixels,
for up to 10,000 iterations.
Render-guided Re-Correspondence (RRC) is applied
to improve the initial estimate,
followed by MegaPose refinement.

\subsection{Comparison with State of the Art}
\label{sec:main_results}

\begin{table}[t]
    \begin{center}
        \resizebox{\textwidth}{!}{%
            \begin{tabular}{l|c|ccccccc|c}
                \hline
                \multirow{2}{*}{Method} &
                \multirow{2}{*}{Refinement} &
                \multicolumn{7}{c|}{Dataset} &
                \multirow{2}{*}{Mean} \\
                \cline{3-9}
                & & LM-O & T-LESS & TUD-L & IC-BIN & ITODD & HB & YCB-V & \\
                \hline\hline

                \multicolumn{10}{l}{%
                \small\textit{Training-free methods (no task-specific training)}} \\
                \hline

                FoundPose~\cite{FoundPose} & - &
                \textbf{39.6} & \underline{33.8} & \underline{46.7} &
                \textbf{23.9} & \underline{20.4} & \underline{50.8} &
                \underline{45.2} & \underline{37.2} \\

                B2TFPose (Ours) & - &
                \underline{38.9} & \textbf{34.6} & \textbf{58.1} &
                \underline{22.8} & \textbf{23.4} & \textbf{58.3} &
                \textbf{48.7} & \textbf{40.7} \\
                \cdashline{1-10}

                FoundPose~\cite{FoundPose} & Feat + MegaPose &
                \underline{55.7} & \underline{51.0} & \underline{63.3} &
                \textbf{43.3} & \underline{35.7} & \underline{69.7} &
                \textbf{66.1} & \underline{55.0} \\

                B2TFPose (Ours) & RRC + MegaPose &
                \textbf{58.6} & \textbf{52.0} & \textbf{69.6} &
                \underline{40.0} & \textbf{35.8} & \textbf{73.1} &
                \textbf{66.1} & \textbf{56.4} \\
                \hline\hline

                \multicolumn{10}{l}{%
                \small\textit{Methods requiring training}} \\
                \hline

                MegaPose~\cite{MegaPose} & - &
                22.9 & 17.7 & 25.8 & 15.2 & 10.8 & 25.1 & 28.1 & 20.8 \\

                GigaPose~\cite{GigaPose} & - &
                \underline{29.9} & \underline{27.3} & \underline{30.2} &
                \underline{23.1} & \underline{18.8} & \underline{34.8} &
                \underline{29.0} & \underline{27.6} \\

                GenFlow~\cite{GenFlow} & - &
                25.0 & 21.5 & 30.0 & 16.8 & 15.4 & 28.3 & 27.7 & 23.5 \\

                Co-op~\cite{coop} & - &
                \textbf{59.7} & \textbf{59.2} & \textbf{64.2} &
                \textbf{45.8} & \textbf{39.1} & \textbf{78.1} &
                \textbf{62.6} & \textbf{58.4} \\
                \cdashline{1-10}

                MegaPose~\cite{MegaPose} & MegaPose &
                49.9 & 47.7 & 65.3 & 36.7 & 31.5 & 65.4 & 60.1 & 50.9 \\

                GigaPose~\cite{GigaPose} & MegaPose &
                \underline{55.6} & \underline{54.6} & 57.8 &
                \underline{44.3} & 37.8 & 69.3 &
                \underline{63.4} & 54.7 \\

                GenFlow~\cite{GenFlow} & GenFlow &
                54.7 & 51.4 & \underline{67.0} & 43.7 & \underline{38.4} &
                \underline{73.0} & 61.9 & \underline{55.7} \\

                Co-op~\cite{coop} & Co-op &
                \textbf{64.2} & \textbf{63.5} & \textbf{71.7} &
                \textbf{51.2} & \textbf{47.3} & \textbf{83.2} &
                \textbf{67.0} & \textbf{64.0} \\
                \hline

            \end{tabular}%
        }
    \end{center}
    \caption{Comparison with state-of-the-art zero-shot RGB pose estimation
        methods on the seven BOP core datasets.
        Values show the BOP Average Recall.
        Methods are grouped by whether they require training on
        object-specific or task-specific data.
        ``Refinement'' indicates use of a dedicated pose refiner.
        Dashed lines separate the \textit{with} and \textit{without}
        refinement sub-groups
        within each category.
        All methods use CNOS~\cite{CNOS} for instance segmentation.
    All results are taken either from the original papers or the BOP leaderboard.}
    \label{tab:main}
\end{table}

Table~\ref{tab:main} compares B2TFPose against recent published zero-shot
RGB methods on the seven BOP core datasets.
Methods are partitioned into two categories: training-free methods
that require no training on object-specific or task-specific data, and
methods requiring additional
task-specific training on synthetic data.
Within each category, results are further split by whether a dedicated
pose refiner is applied.
All results for competing methods are taken from their respective
publications or from the BOP leaderboard under identical evaluation
conditions.
We use the single-hypothesis setting where methods report results for
both single-hypothesis and multi-hypothesis modes, matching the
operating mode of
B2TFPose, which produces one pose estimate per object instance.
The multi-mask selection over up to two CNOS candidates
(Sec.~\ref{para:mask-selection}) addresses segmentation ambiguity, not pose
hypothesis diversity, and is not equivalent to multi-hypothesis pose
sampling.

\noindent\textbf{Accuracy without refinement.}
Among training-free methods without a dedicated refiner, B2TFPose
achieves 40.7 mean AR, establishing a new state of the art in this
category.
FoundPose~\cite{FoundPose}, the closest comparable baseline, which
also builds on frozen visual features and PnP without any task-specific
training, reaches 37.2 mean AR, a gap of 3.5 points.
B2TFPose leads by substantial margins on TUD-L ($+$11.4), HB ($+$7.5),
YCB-V ($+$3.5), ITODD ($+$3.0), and T-LESS ($+$0.8).
The advantage is most pronounced on datasets where broad viewpoint
coverage and in-plane rotation diversity are decisive: our 1944 template
pool, selected with a geodesic diversity constraint, provides
geometrically complementary correspondences that FoundPose's smaller
template set cannot match.
FoundPose retains a narrow advantage on LM-O ($-$0.7) and IC-BIN
($-$1.1), datasets dominated by heavy occlusion and dense clutter, where
expanded viewpoint coverage is less decisive than occlusion-robust
features.
Co-op~\cite{coop} achieves a substantially higher 58.4 mean AR, but this
accuracy comes at the cost of training on large-scale synthetic datasets,
placing it outside the training-free category.

\noindent\textbf{Accuracy with refinement.}
With Render-guided Re-Correspondence (RRC) followed by MegaPose,
B2TFPose reaches 56.4 mean AR, again the best result among training-free
methods.
FoundPose, refined with featuremetric alignment and MegaPose, reaches
55.0 AR and places second in this category.
B2TFPose leads FoundPose (refined) on TUD-L ($+$6.3), HB ($+$3.4),
LM-O ($+$2.9), T-LESS ($+$1.0), and ITODD ($+$0.1).
FoundPose retains an advantage on IC-BIN ($-$3.3); both methods are
tied on YCB-V (66.1).
Among trained methods with refinement, GenFlow reaches 55.7 AR and
GigaPose with MegaPose reaches 54.7 AR, both falling below B2TFPose.
Co-op leads at 64.0 AR and is the only trained method to exceed ours,
while requiring training on large-scale synthetic data.
Qualitative results are provided in the supplementary material.
\begin{figure}[t]
    \centering
    \includegraphics[width=0.80\linewidth]{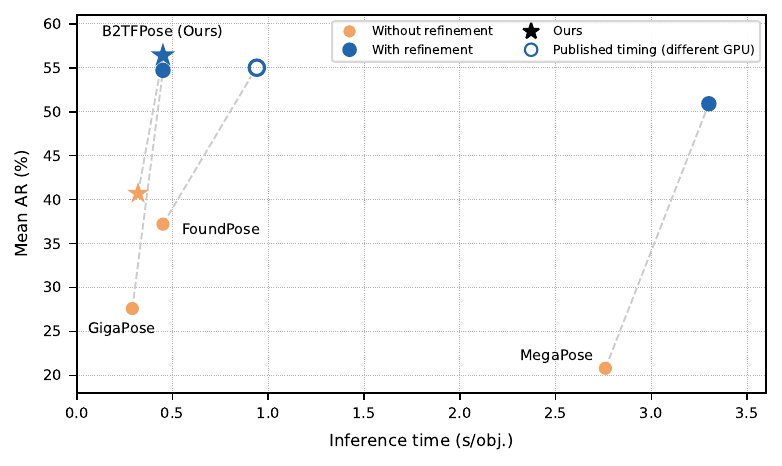}
    \vspace{2pt}
    \caption{Speed-accuracy trade-off across the seven BOP datasets.
        \textbf{Orange} markers denote performance without pose refinement;
        \textbf{blue} markers denote performance with pose refinement.
        Dashed lines connect the two conditions for the same method.
        The star~($\bigstar$) marks our method (B2TFPose).
        Inference time is measured per object per image on TUD-L.
        The hollow marker for FoundPose (with refinement) indicates that
        the reported time
        is taken from the BOP leaderboard and was measured on
        different hardware,
        as its
        full-pipeline code is not publicly released.
        Co-op~\cite{coop} and GenFlow~\cite{GenFlow} are excluded as their
        code is not publicly released.
    Upper-left is better.}
    \label{fig:speed}
\end{figure}

\noindent\textbf{Speed.} Figure~\ref{fig:speed} plots the
speed-accuracy trade-off on TUD-L
(seconds per object instance per image), measured by running each
method's released code on our hardware (NVIDIA RTX~3090).
Co-op~\cite{coop} and GenFlow~\cite{GenFlow} are excluded as their
code is not publicly released.

Without refinement, B2TFPose runs in 0.32\,s.
Only GigaPose (0.29\,s) is marginally quicker, but at substantially lower
accuracy (27.6 vs.\ 40.7 AR, a gap of 13.1 AR points).
FoundPose (0.45\,s) is slower and 3.5 AR points behind.
B2TFPose therefore occupies the optimal training-free position: highest
accuracy at competitive speed.

With refinement, B2TFPose reaches 56.4 AR in 0.45\,s, matching
GigaPose (0.45\,s) in speed while exceeding it by 1.7 AR points.
Compared to FoundPose with refinement, whose published timing of 0.94\,s
was measured on a comparable P100 GPU (shown with a hollow blue circle
marker in Fig.~\ref{fig:speed}),
B2TFPose achieves a 1.4 AR point accuracy advantage over FoundPose with
refinement, at less than half its published inference time.
In both settings, B2TFPose is the training-free method closest to the
upper-left corner of the speed-accuracy plot.

\subsection{Ablation Studies and Hyperparameter Tuning}
\label{sec:ablation}

The RRC ablation study is conducted on all 7 BOP datasets, whereas
the remaining ablation and hyperparameter studies are conducted on
TUD-L, which offers a representative mix of object geometries and
illumination conditions; the backbone ablation additionally includes
IC-BIN, the one dataset on which FoundPose leads, to test whether its
effect generalizes beyond TUD-L.
Unless otherwise stated, the base configuration uses $k\!=\!125$ templates,
3 patch neighbors, and a PnP reprojection threshold of 4 pixels.

\begin{sloppypar}
    \noindent\textbf{Render-guided Re-Correspondence (RRC).}
    RRC is one of the novel components of B2TFPose (Sec.~\ref{sec:method-rrc}) and
    Table~\ref{tab:ablation_rrc} isolates its contribution.
    RRC alone (row 2) yields gains on occlusion-heavy and textureless
    datasets where the initial PnP estimate is most uncertain, including
    LM-O ($+$4.3), T-LESS ($+$2.4), ITODD ($+$1.8), and YCB-V ($+$0.8).
    On TUD-L, IC-BIN, and HB, RRC alone produces small performance
    degradations ($-$0.8, $-$0.3, and $-$1.3 respectively) which
    we attribute to the fact that
    when the first-round pose is near-correct, re-projecting it occasionally
    displaces a good solution.
    Applying MegaPose without RRC (row 3) yields the dominant share of the
    accuracy gain, advancing mean AR from 40.7 to 55.5 and confirming that
    the iterative render-and-compare refiner makes a significant contribution to
    the final accuracy.
    The full pipeline RRC+MegaPose (row 4) reaches 56.4 mean AR. On top of
    MegaPose alone, RRC provides gains on LM-O ($+$2.4),
    T-LESS ($+$2.0), ITODD ($+$2.2), and YCB-V ($+$1.0), confirming that a
    tighter initial hypothesis steers the iterative refiner away from local
    optima on textureless and occlusion-heavy datasets.
    Small accuracy drops on IC-BIN ($-$0.5) and HB ($-$0.6) mirror the pattern
    seen for RRC alone.
    RRC and MegaPose are therefore complementary; RRC acts as a
    test-time correspondence bridge, not merely as a pre-processor for the refiner.
    Visual illustration of the RRC step and its effect on the pose estimate is
    provided in the supplementary material.
\end{sloppypar}

\begin{table}[t]
    \begin{center}
        \resizebox{\textwidth}{!}{%
            \begin{tabular}{c|c|c|ccccccc|c}
                \hline
                \multirow{2}{*}{\#} &
                \multirow{2}{*}{RRC} &
                \multirow{2}{*}{MegaPose} &
                \multicolumn{7}{c|}{Dataset} &
                \multirow{2}{*}{Mean} \\
                \cline{4-10}
                & & & LM-O & T-LESS & TUD-L & IC-BIN & ITODD & HB & YCB-V & \\
                \hline\hline

                1 & \ding{55} & \ding{55} &
                38.9 & 34.6 & 58.1 & 22.8 & 23.4 & 58.3 & 48.7 & 40.7\\

                2 & \ding{51} & \ding{55} &
                43.2 & 37.0 & 57.3 & 22.5 & 25.2 & 57.0 & 49.5 & 41.7\\

                3 & \ding{55} & \ding{51} &
                56.2 & 50.0 & \textbf{69.6} & \textbf{40.5} & 33.6 &
                \textbf{73.7} & 65.1 & 55.5\\

                4 & \ding{51} & \ding{51} &
                \textbf{58.6} & \textbf{52.0} & \textbf{69.6} & 40.0 &
                \textbf{35.8} & 73.1 & \textbf{66.1} & \textbf{56.4}\\
                \hline

        \end{tabular}}
    \end{center}
    \caption{Ablation of the RRC and MegaPose refinement stages.
        Row~1 is the base pipeline with no refinement; rows~2-4 add RRC alone,
        MegaPose alone, and both in combination. AR~(\%) is reported per dataset
    and as the mean across all seven BOP core datasets.}
    \label{tab:ablation_rrc}
\end{table}

\noindent\textbf{Effect of Backbone Choice.}
To test whether B2TFPose's accuracy relative to FoundPose~\cite{FoundPose}
(which uses DINOv2~\cite{DINOv2}) stems from our retrieval and
refinement design or from a stronger backbone, we swap only the
backbone for DINOv2 and re-run on TUD-L (our largest margin over
FoundPose) and IC-BIN (where FoundPose leads).
With PnP-only/with
RRC/with
full refinement,
AR is 57.3/56.3/68.1 on TUD-L ($-0.8/-1.0/-1.5$ vs.\
DINOv3) and 23.1/26.5/42.0 on IC-BIN ($+0.3/+4.0/+2.0$). The
opposite-direction effect, a drop on TUD-L but a gain on IC-BIN,
shows the reported gains are not simply inherited from DINOv3's
representational strength.

\begin{figure}[t]
    \centering
    \includegraphics[width=\linewidth]{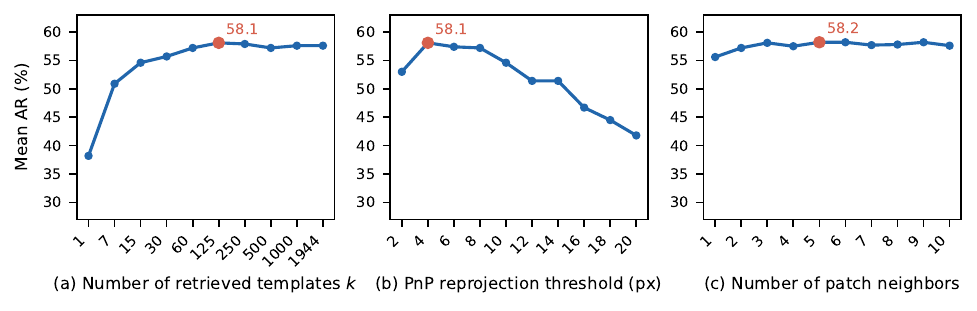}
    \vspace{2pt}
    \caption{Effect of three key hyperparameters on TUD-L AR.
        \textbf{(a)}~Number of retrieved templates $k$: performance peaks
        at $k\!=\!125$;
        using all 1944 templates is marginally worse as the geodesic
        diversity constraint
        loses force for large $k$.
        \textbf{(b)}~PnP reprojection threshold: the optimal is 4\,px;
        tighter thresholds
        reject valid correspondences while looser ones admit outlier
        correspondences.
        \textbf{(c)}~Number of patch neighbors: accuracy saturates at 3 neighbors
        and remains stable beyond; 3 is adopted as the default as values of 5 and
    above reach only 0.1~AR higher at greater retrieval cost.}
    \label{fig:ablations}
\end{figure}

\noindent\textbf{Number of Retrieved Templates ($k$).}
Figure~\ref{fig:ablations}(a) visualizes TUD-L accuracy as the number of
retrieved templates varies from 1 to 1944 (all templates).
Performance rises steeply from $k\!=\!1$ (38.2 AR) to $k\!=\!125$
(58.1 AR), then plateaus and slightly declines for larger values.
Using all 1944 templates (57.6 AR) is marginally worse than
$k\!=\!125$.
We attribute this to the geodesic diversity constraint: with
$k\!=\!125$, non-maximum suppression enforces a minimum $20^\circ$
inter-view separation, yielding a set of templates that spans the
viewpoint sphere with minimal redundancy.
Increasing $k$ beyond 125 admits additional templates from viewpoints
already well-covered, introducing redundant correspondences that
slightly dilute the RANSAC inlier pool.
We therefore set $k\!=\!125$ as our default.

\noindent\textbf{PnP Reprojection Threshold.}
Figure~\ref{fig:ablations}(b) shows TUD-L accuracy as the RANSAC+PnP
inlier threshold varies from 2 to 20 pixels.
The optimal threshold is 4 pixels (58.1 AR), with performance
declining sharply on both sides.
At 2 pixels, the constraint is strict, rejecting too many
valid correspondences and degrading RANSAC robustness (53.0 AR,
$-$5.1 from the optimum).
Above 4 pixels, accuracy declines monotonically, moving from 57.2 AR at
8\,px to 41.8 AR at 20\,px, as an increasing fraction of outlier
correspondences pass the inlier check and contaminate the PnP
pool.
We adopt 4 pixels as the default reprojection threshold.

\noindent\textbf{Number of Patch Neighbors.}
Figure~\ref{fig:ablations}(c) reports the effect of varying the
number of nearest template patch descriptors retrieved per query patch
before MNN filtering.
Accuracy rises from 55.6 AR at 1 neighbor to 58.1 AR at 3 neighbors,
then saturates; values of 5, 6, and 9 neighbors each reach 58.2 AR,
only 0.1 above the default.
A single neighbor provides too few candidates for MNN filtering to
reject ambiguous matches effectively; beyond 3, additional neighbors
yield diminishing returns as the bidirectional consistency check
already discards a significant number of outlier matches.
We adopt 3 neighbors as the default, as it achieves near-optimal
accuracy at the lowest retrieval cost.

\noindent\textbf{Effect of Geodesic Diversity in Retrieval.}
To isolate the contribution of the geodesic diversity constraint itself, we
disable it and retrieve the raw top-$k\!=\!125$ templates by cosine
similarity alone, on TUD-L; this reduces AR by 0.6/2.3/2.8 at the
PnP-only stage, with RRC, and with full refinement, relative to the
diversity-enforcing default. Skipping retrieval altogether, using all
162 base viewpoints as the template set matched against every query
patch rather than a similarity-selected subset, reduces AR by
2.6/2.3/2.6 at the same three stages. Both results confirm that retrieval
and its diversity
constraint contribute independently rather than being redundant with
downstream correspondence matching or PnP's own robustness to
outliers.

\section{Conclusion}
We presented B2TFPose, a training-free zero-shot 6DoF pose estimation
method for novel objects from RGB images, relying solely on a frozen
DINOv3 vision transformer within the pose estimation pipeline, without
any object-specific or task-specific training.
Three contributions drive its design: geodesic non-maximum suppression
for viewpoint-diverse template retrieval, Render-guided Re-Correspondence
(RRC) for test-time correspondence refinement anchored at the initial
pose estimate, and multi-mask hypothesis selection to resolve segmentation
ambiguity.
On the seven core datasets of the BOP Benchmark, B2TFPose establishes
state-of-the-art performance among training-free RGB methods,
surpassing FoundPose~\cite{FoundPose} in both with and without refinement
settings and outperforming
trained methods including GigaPose~\cite{GigaPose} and GenFlow~\cite{GenFlow}
despite requiring no training data, at competitive inference speed.
The method requires a 3D CAD model at test time, limiting applicability
to settings where such models are unavailable, and performance degrades
under heavy occlusion and dense clutter where appearance-based matching
without depth cues is less robust.
We hope these results encourage further exploration of classical local
feature matching through large-scale pretrained vision models as a
generalizable and efficient paradigm for 6DoF pose estimation.

\bibliographystyle{unsrtnat}
\bibliography{references}

\newpage
\makeatletter
\vbox{%
    \hsize\textwidth
    \linewidth\hsize
    \vskip 0.1in
    \@toptitlebar
    \centering
    {\LARGE\sc Supplementary Material for:\\[4pt]
    Back to the Feature: Zero-Shot 6DoF Pose Estimation via Dense Local Features\par}
    \@bottomtitlebar
}
\makeatother

\appendix
\section{Overview}
This supplementary material provides qualitative results in support of our
method, B2TFPose,
on the BOP benchmark datasets, and a visual explanation of the
Render-guided Re-Correspondence (RRC) step.
Specifically, this document contains:
\begin{itemize}
    \item \textbf{Section~\ref{sec:qual} - Qualitative Results:}
        Representative success and failure cases on BOP datasets.
        Each row shows a query image alongside pose overlays at the
        estimated and ground-truth poses.
    \item \textbf{Section~\ref{sec:rrc} - RRC Visualization:}
        Three cases illustrating how Render-guided Re-Correspondence (RRC)
        improves the initial RANSAC+PnP estimate before MegaPose refinement,
        with the ten object-specific renders generated at the initial pose
        shown for each case.
\end{itemize}

\section{Qualitative Results}
\label{sec:qual}

\begin{figure}[H]
    \centering
    \setlength{\tabcolsep}{2pt}
    \begin{tabular}{cccc}
        \small Query &
        \small No refinement &
        \small RRC + MegaPose &
        \small Ground truth \\[2pt]

        \includegraphics[width=\imgwidth]{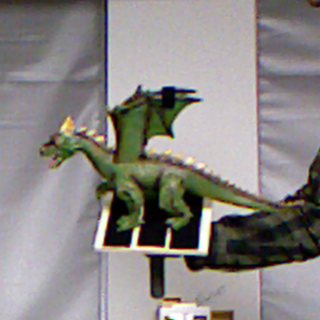}
        &
        \includegraphics[width=\imgwidth]{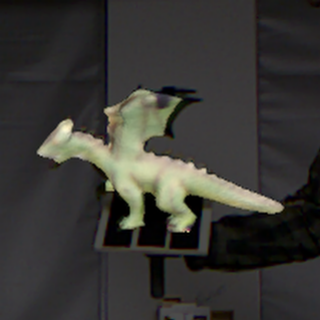}
        &
        \includegraphics[width=\imgwidth]{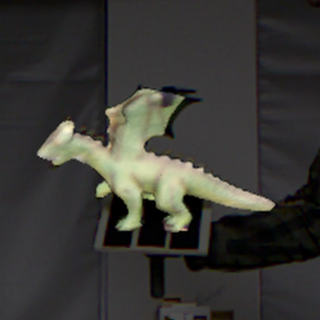}
        &
        \includegraphics[width=\imgwidth]{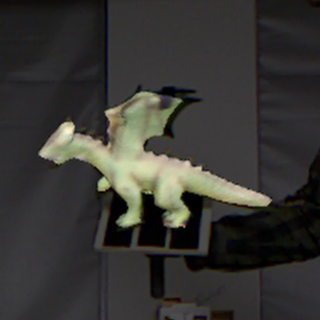}
        \\
        \multicolumn{4}{c}{\footnotesize\textit{(a) Accurate pose estimation both
        without and with refinement (TUD-L).}} \\[4pt]

        \includegraphics[width=\imgwidth]{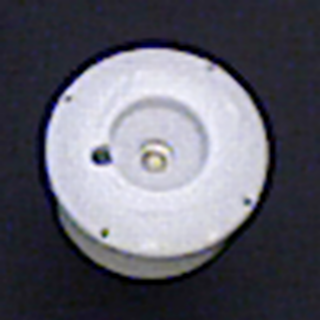}
        &
        \includegraphics[width=\imgwidth]{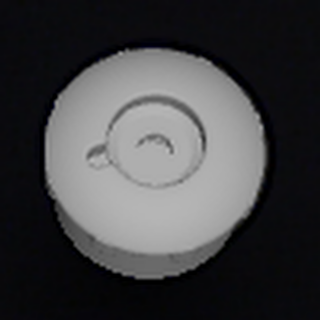}
        &
        \includegraphics[width=\imgwidth]{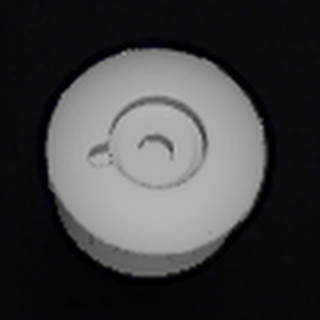}
        &
        \includegraphics[width=\imgwidth]{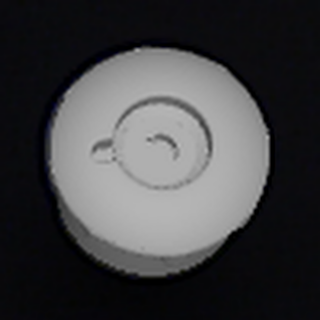}
        \\
        \multicolumn{4}{c}{\parbox{0.8\linewidth}{\centering\footnotesize\textit{(b)
                    Accurate pose
                    estimation without and with refinement on a \\near-textureless
        object (T-LESS).}}} \\[8pt]

        \includegraphics[width=\imgwidth]{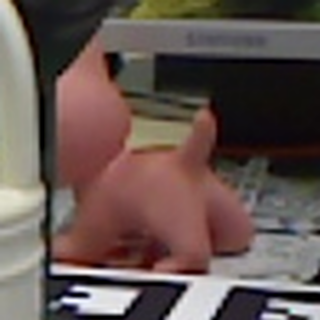}
        &
        \includegraphics[width=\imgwidth]{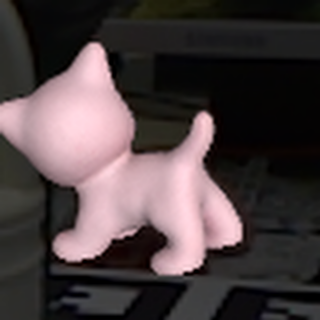}
        &
        \includegraphics[width=\imgwidth]{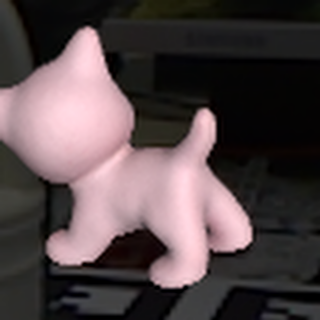}
        &
        \includegraphics[width=\imgwidth]{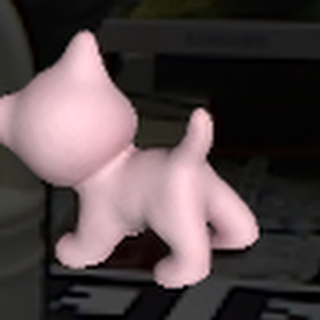}
        \\
        \multicolumn{4}{c}{\footnotesize\textit{(c) Accurate pose estimation on a
        partially occluded object (LM-O).}} \\[4pt]

        \includegraphics[width=\imgwidth]{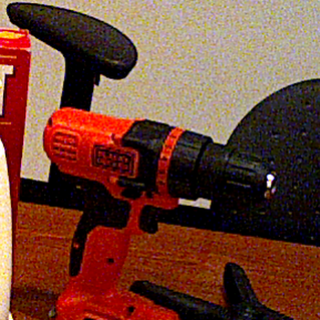}
        &
        \includegraphics[width=\imgwidth]{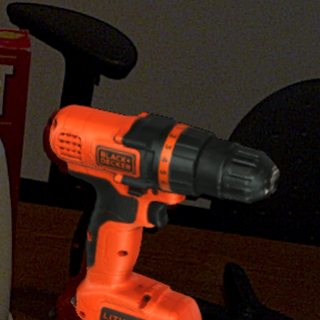}
        &
        \includegraphics[width=\imgwidth]{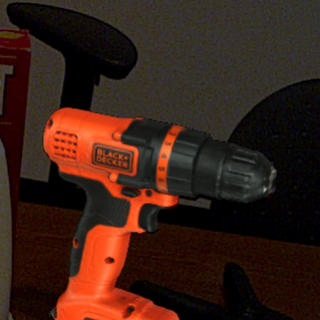}
        &
        \includegraphics[width=\imgwidth]{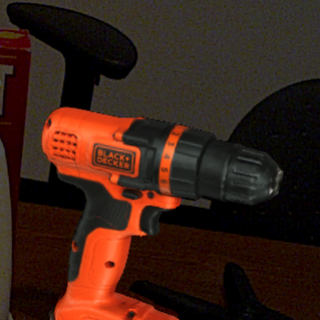}
        \\
        \multicolumn{4}{c}{\footnotesize\textit{(d) Refinement tightens a slightly
        off initial estimate to the correct pose (YCB-V).}} \\[4pt]

        \includegraphics[width=\imgwidth]{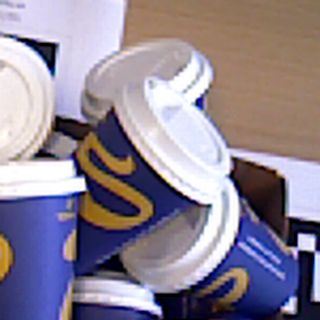}
        &
        \includegraphics[width=\imgwidth]{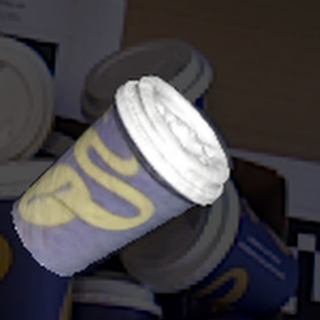}
        &
        \includegraphics[width=\imgwidth]{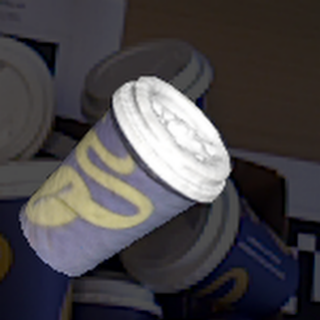}
        &
        \includegraphics[width=\imgwidth]{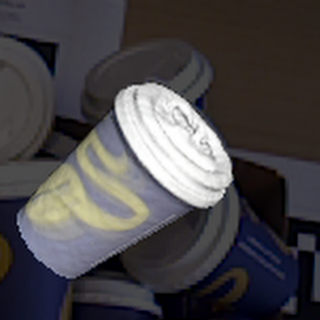}
        \\
        \multicolumn{4}{c}{\parbox{0.8\linewidth}{\centering\footnotesize\textit{(e)
                    Refinement closes a small
                    gap in the pose estimation in a bin-picking scenario
                    \\with mutual occlusion and a symmetric
        object (IC-BIN).}}} \\

    \end{tabular}
    \vspace{12pt}
    \caption{{B2TFPose qualitative results.
            The 3D object model is overlaid at the
            estimated pose. The method achieves accurate pose estimates across
            diverse object geometries and scene conditions; when the initial
            estimate is slightly off, refinement recovers the
    correct pose.}}
    \label{fig:qual_success}
\end{figure}

\begin{figure}[H]
    \centering
    \setlength{\tabcolsep}{2pt}
    \begin{tabular}{cccc}
        \small Query &
        \small No refinement &
        \small RRC + MegaPose &
        \small Ground truth \\[2pt]

        \includegraphics[width=\imgwidth]{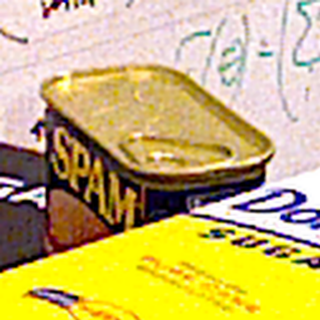}
        &
        \includegraphics[width=\imgwidth]{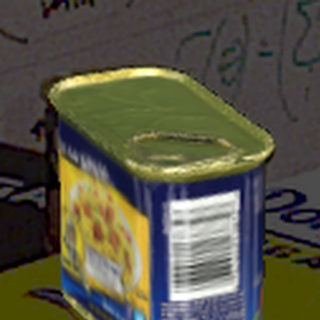}
        &
        \includegraphics[width=\imgwidth]{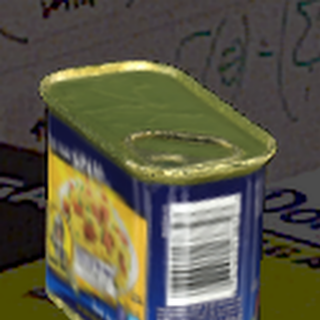}
        &
        \includegraphics[width=\imgwidth]{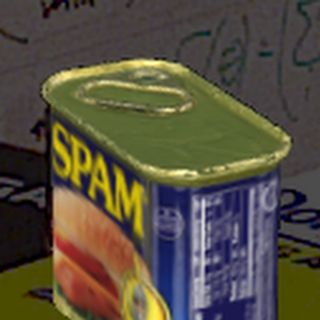}
        \\
        \multicolumn{4}{c}{\parbox{0.8\linewidth}{\centering\footnotesize\textit{(a)
                    Rotational symmetry
                    and partial occlusion: \\patch ambiguity and model noise lead to
        an incorrect estimate.}}} \\[6pt]

        \includegraphics[width=\imgwidth]{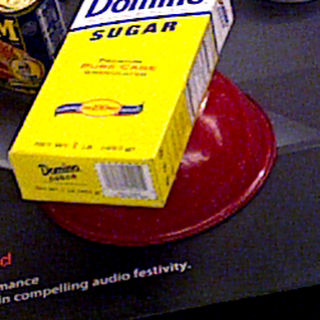}
        &
        \includegraphics[width=\imgwidth]{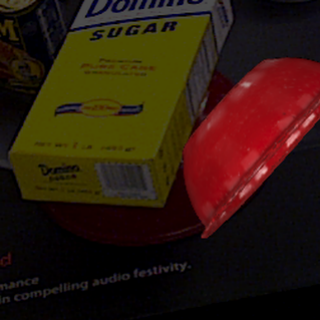}
        &
        \includegraphics[width=\imgwidth]{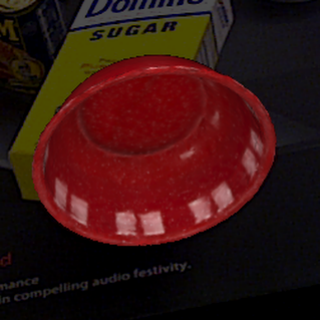}
        &
        \includegraphics[width=\imgwidth]{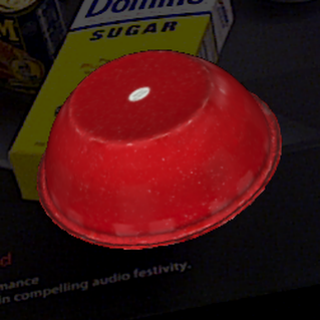}
        \\
        \multicolumn{4}{c}{\footnotesize\textit{(b) Severe occlusion: too few
        foreground patches survive for reliable correspondence.}} \\[4pt]

        \includegraphics[width=\imgwidth]{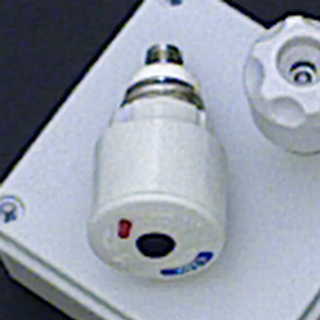}
        &
        \includegraphics[width=\imgwidth]{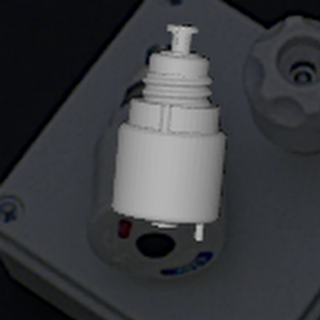}
        &
        \includegraphics[width=\imgwidth]{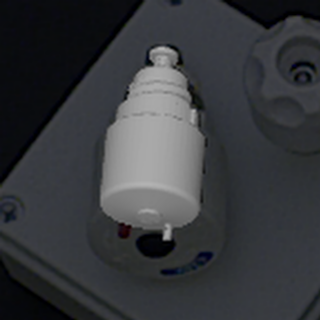}
        &
        \includegraphics[width=\imgwidth]{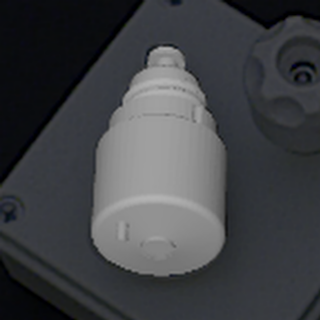}
        \\
        \multicolumn{4}{c}{\footnotesize\textit{(c) Near-textureless object:
        patch features lack discriminative detail.}} \\[4pt]

        \includegraphics[width=\imgwidth]{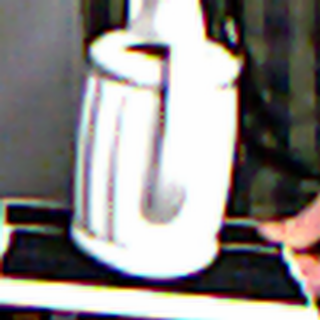}
        &
        \includegraphics[width=\imgwidth]{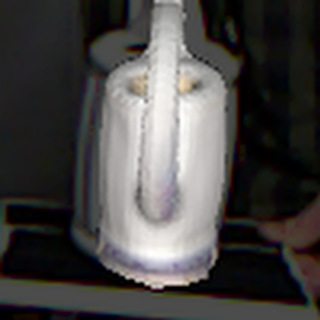}
        &
        \includegraphics[width=\imgwidth]{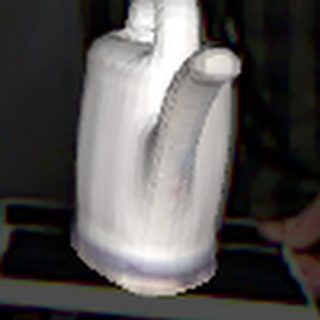}
        &
        \includegraphics[width=\imgwidth]{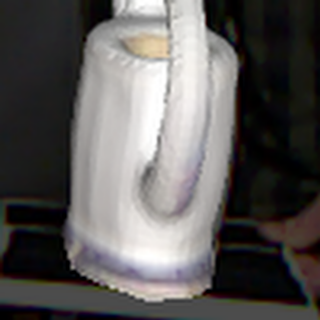}
        \\
        \multicolumn{4}{c}{\footnotesize\textit{(d) Challenging lighting / motion
        blur: feature extraction is unreliable.}} \\[4pt]

        \includegraphics[width=\imgwidth]{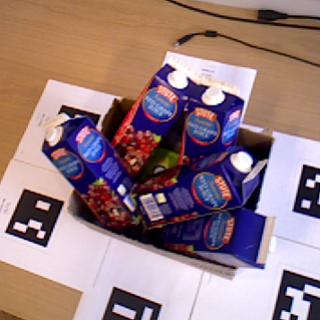}
        &
        \includegraphics[width=\imgwidth]{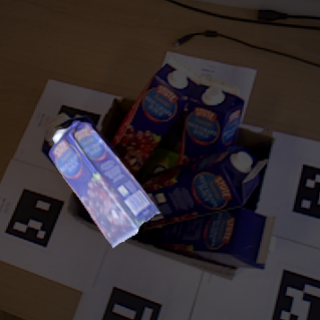}
        &
        \includegraphics[width=\imgwidth]{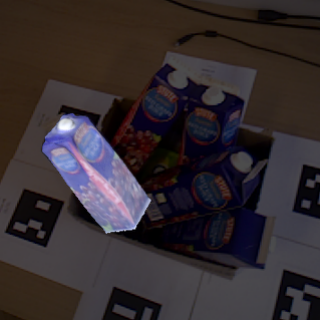}
        &
        \includegraphics[width=\imgwidth]{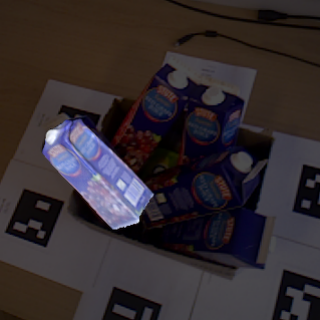}
        \\
        \multicolumn{4}{c}{\parbox{0.8\linewidth}{\centering\footnotesize\textit{(e)
                    Bin-picking with
                    multiple identical instances: \\inter-object ambiguity causes
        estimates to fail (IC-BIN).}}} \\

    \end{tabular}
    \vspace{12pt}
    \caption{{B2TFPose failure cases. B2TFPose degrades under rotational
            symmetry, severe occlusion, textureless appearance, and
            inter-object ambiguity in bin-picking scenarios; inherent
            limitations of appearance-based matching without explicit
    depth cues.}}
    \label{fig:qual_failure}
\end{figure}

Figures~\ref{fig:qual_success} and~\ref{fig:qual_failure} show
qualitative results of B2TFPose across five BOP datasets (TUD-L,
T-LESS, LM-O, YCB-V, IC-BIN). The other two datasets (HB, ITODD)
were excluded as the ground-truth poses are not publicly available.
Each row shows a single object instance: the query RGB crop, the estimated
pose without refinement, the estimated pose after RRC and MegaPose
refinement, and the ground-truth pose, all visualised as an
overlay of the 3D object model projected onto the image.
A brief description below each row summarises the scenario and outcome.

\section{Render-guided Re-Correspondence (RRC) Visualization}
\label{sec:rrc}

\begin{figure}[H]
    \centering
    \setlength{\tabcolsep}{2pt}
    \begin{tabularx}{\linewidth}{@{}*{5}{>{\centering\arraybackslash}X}@{}}
        \small Query &
        \small Initial &
        \small Initial+MegaPose &
        \small Initial+RRC &
        \small Initial+RRC+MP \\[2pt]

        \includegraphics[width=\linewidth]{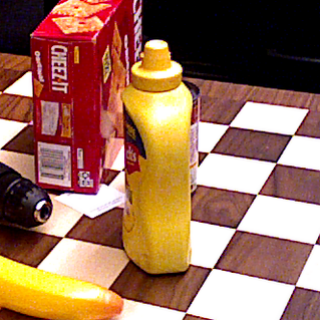}
        &
        \includegraphics[width=\linewidth]{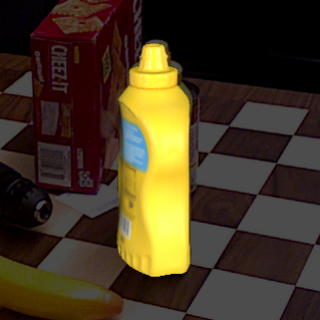}
        &
        \includegraphics[width=\linewidth]{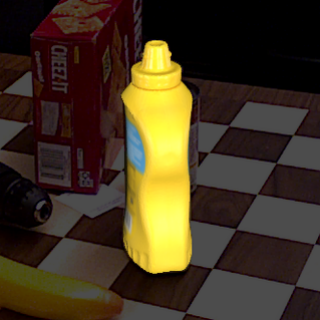}
        &
        \includegraphics[width=\linewidth]{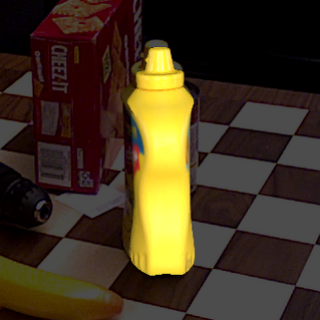}
        &
        \includegraphics[width=\linewidth]{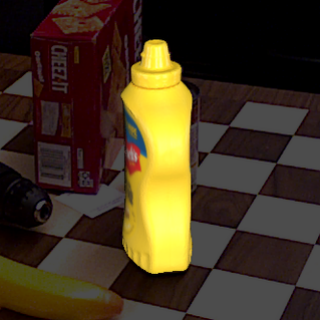}
        \\
        \multicolumn{5}{@{}c@{}}{%
        \includegraphics[width=\linewidth]{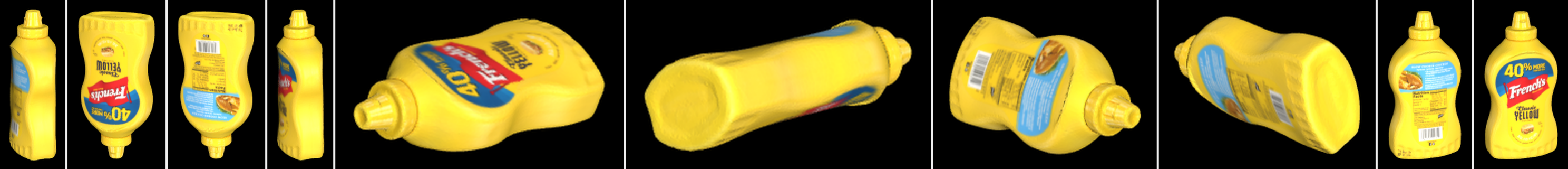}}
        \\[12pt]

        \includegraphics[width=\linewidth]{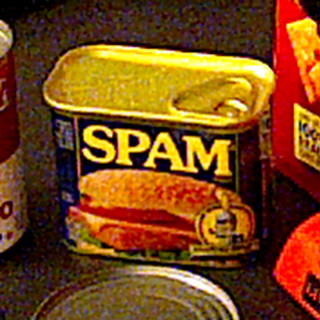}
        &
        \includegraphics[width=\linewidth]{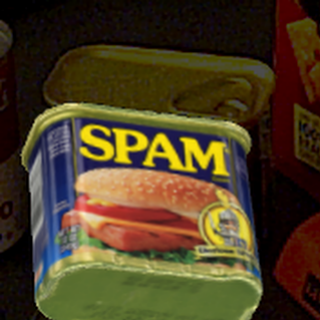}
        &
        \includegraphics[width=\linewidth]{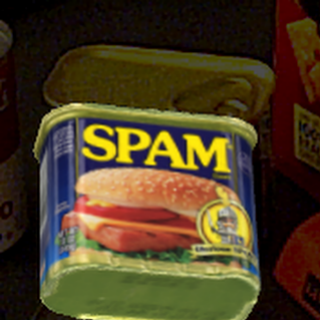}
        &
        \includegraphics[width=\linewidth]{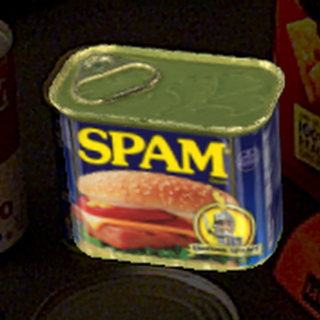}
        &
        \includegraphics[width=\linewidth]{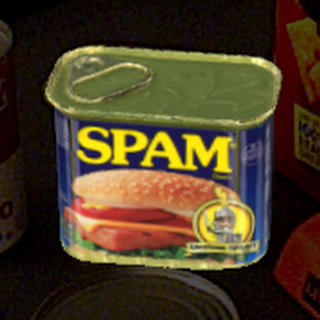}
        \\

        \multicolumn{5}{@{}c@{}}{%
        \includegraphics[width=\linewidth]{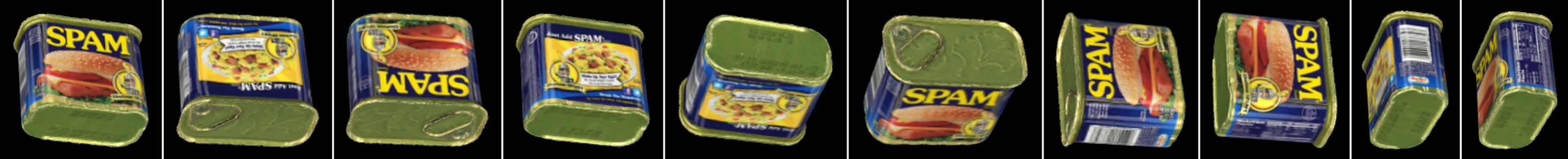}}
        \\[12pt]

        \includegraphics[width=\linewidth]{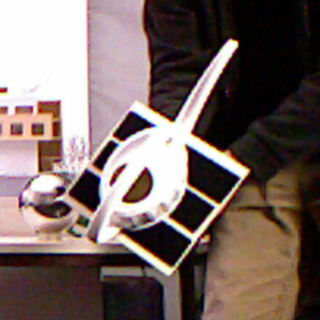}
        &
        \includegraphics[width=\linewidth]{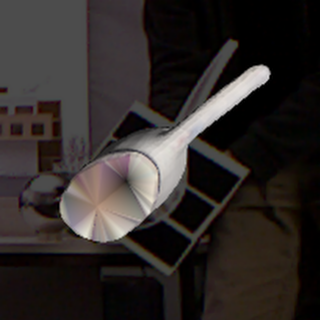}
        &
        \includegraphics[width=\linewidth]{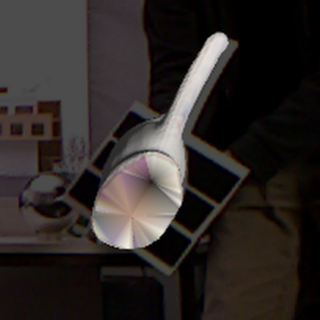}
        &
        \includegraphics[width=\linewidth]{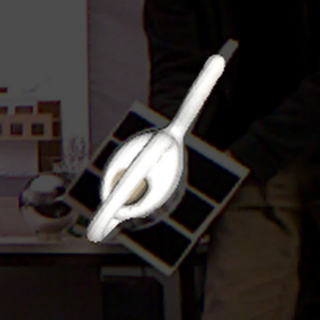}
        &
        \includegraphics[width=\linewidth]{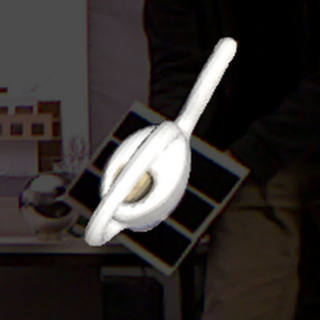}
        \\

        \multicolumn{5}{@{}c@{}}{%
        \includegraphics[width=\linewidth]{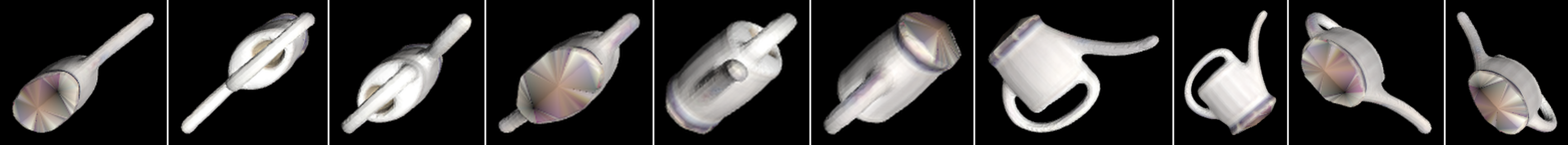}}
        \\

    \end{tabularx}
    \vspace{15pt}
    \caption{RRC visualization for three object instances.
        \textbf{Top row of each case:} query image and pose overlays at four
        pipeline stages.
        \textbf{Bottom row of each case:} the ten pose variants rendered at the
        initial estimate and used by RRC for a second round of 2D-3D
    correspondences.}
    \label{fig:rrc_viz}
\end{figure}

Figure~\ref{fig:rrc_viz} illustrates three cases where RRC improves pose
accuracy.
Each case has two rows: the top row shows the query image and pose overlays
at four pipeline stages: (1)~the initial RANSAC+PnP estimate;
(2)~MegaPose applied directly to that estimate (illustrating how MegaPose
can converge to a local optimum when the initialization is coarse);
(3)~RRC applied to the initial estimate without MegaPose (showing how RRC
produces a tighter hypothesis); (4)~the full RRC+MegaPose pipeline
(showing that the tighter RRC initialization guides MegaPose to the
correct pose, compared to column~2).
The 3D object model is projected onto the query image at each estimated pose.
The bottom row shows the ten object-specific views rendered at the initial
estimate that RRC uses to establish a second round of 2D-3D correspondences;
these renders are significantly closer in appearance to the query than the
offline template library, enabling a tighter RANSAC+PnP solve before the
final MegaPose refinement.

\end{document}